\documentclass[twoside,leqno,twocolumn]{article}

\usepackage[letterpaper]{geometry}

\usepackage{ltexpprt}
\usepackage{hyperref}
\usepackage{comment}
\usepackage{lipsum}
\usepackage{rotating} 
\usepackage{geometry}
\usepackage{amsmath}
\usepackage{amssymb}
\usepackage{wasysym}
\usepackage{algorithm}
\usepackage{algorithmic}
\usepackage{graphics}
\usepackage{graphicx}
\usepackage{float}
\usepackage{xspace}
\usepackage{xcolor}
\usepackage{caption}
\usepackage{fancyhdr}
\usepackage{booktabs}
\usepackage{multirow}
\usepackage[section]{placeins}
\usepackage[T1]{fontenc}
\usepackage{cuted}

\newcommand{\ours}{\textsc{GraphDC}\xspace}

\begin{document}

\newcommand\relatedversion{}
\renewcommand\relatedversion{\thanks{The full version of the paper can be accessed at \protect\url{https://arxiv.org/abs/1902.09310}}} 

\title{\Large \ours: A Divide-and-Conquer Multi-Agent System for Scalable Graph Algorithm Reasoning}
\author{Wenjin Li\thanks{Department of Computer Science, Virginia Tech, Email: \{wenjinl25,jiamingcui\}@vt.edu.}
\and Jiaming Cui$^*$}
\date{}

\maketitle







\begin{abstract}
Large Language Models (LLMs) have demonstrated strong potential for many mathematical problems. However, their performance on graph algorithmic tasks is still unsatisfying, since graphs are naturally more complex in topology and often require systematic multi-step reasoning, especially on larger graphs. Motivated by this gap, we propose \ours, a Divide-and-Conquer multi-agent framework for scalable graph algorithm reasoning. Specifically, inspired by Divide-and-Conquer design, \ours decomposes an input graph into smaller subgraphs, assigns each subgraph to a specialized agent for local reasoning, and uses a master agent to integrate the local outputs with inter-subgraph information to produce the final solution. This hierarchical design reduces the reasoning burden on individual agents, alleviates computational bottlenecks, and improves robustness on large graph instances. Extensive experiments show that \ours consistently outperforms existing methods on graph algorithm reasoning across diverse tasks and scales, especially on larger instances where direct end-to-end reasoning is less reliable.
\end{abstract}


\section{Introduction}\label{sec:introduction}

Graph is a fundamental mathematical concept for representing relationships between entities~\cite{r23,r24,anand2024h2abm,cui2024modeling,cui2025identifying}. Many real-world problems such as infectious disease modeling~\cite{cui2026bridging,venkatesh2026physicsagentabm,chopra2023differentiable}, supply chain optimization~\cite{yaakoubi2022learning}, and urban infrastructure surveillance~\cite{tabassum2021actionable,tabassum2021efficient} can be naturally modeled as graphs, where the objective is to reason over structural dependencies among entities rather than treat instances independently. Therefore, graph algorithm reasoning has become central to numerous domains, including social network analysis~\cite{r6,r7}, bioinformatics~\cite{r44}, and knowledge graph inference~\cite{r8,r9,r10}. Over the years, a wide range of specialized architectures have been proposed~\cite{r26,r27} to tackle these tasks. While these models achieve strong performance, they often suffer from limited generalization and usability~\cite{r28,r29}: achieving state-of-the-art results typically requires task-specific designs such as tailored preprocessing pipelines and decoders, which make them less flexible and harder to adapt. This limits their applicability when task formulations, graph characteristics, or supervision settings vary across datasets and applications.

Recently, large language models (LLMs) have shown impressive capabilities in natural language interaction, interpretability, and generalization across diverse mathematical tasks~\cite{r11,r12,r13,r14,r15,liu2024time,datta2025improving}. These properties make LLMs an appealing alternative to specialized graph models, as they offer a unified interface for solving different reasoning tasks with minimal task-specific redesign. This has sparked growing interest in their potential for graph algorithm reasoning. However, the reasoning capacity of a single LLM is inherently constrained: as graph size and density increase, LLM accuracy on graph algorithm reasoning tasks declines sharply. This limitation is particularly pronounced for algorithmic reasoning, where the model must simultaneously track multi-hop dependencies, preserve structural consistency, and execute multi-step logical operations. As a result, directly applying a single LLM to increasingly large graphs is difficult to scale.

\begin{figure*}[t]
  \centering
  \includegraphics[width=\linewidth]{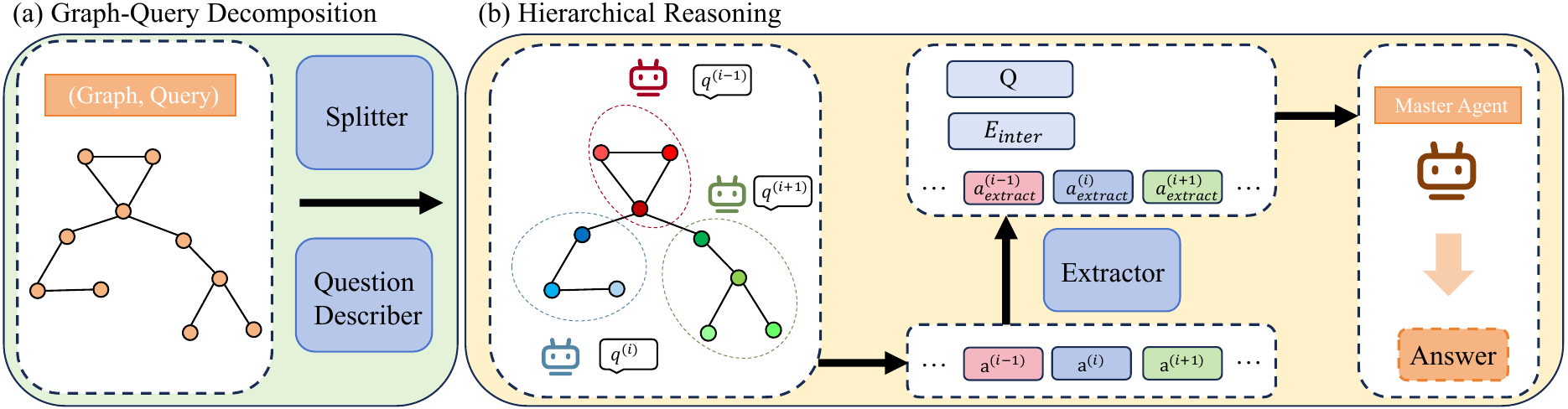}
  \caption{The framework of \ours. Given a mathematical problem on graphs, \ours solves it in two steps. In the \textbf{Graph-Query Decomposition} step, a graph splitter first decomposes the graph into several subgraphs. Each specialized agent then processes one subgraph using a sub-query generated by the question describer and performs intra-subgraph algorithm reasoning. Next, in the \textbf{Hierarchical Reasoning} step, a master agent aggregates the answers from all agents, together with inter-subgraph information, to respond to the original question.}
  \label{fig:figure1}
\end{figure*}

To mitigate this degradation, several strategies have been explored~\cite{r16,r17,r18,r19}. For example, GITA~\cite{r3} augments prompts with graph visualizations, while GraphAgent-Reasoner~\cite{r2} introduces a multi-agent system where a master agent coordinates a set of smaller agents. These studies suggest that adding structure to the reasoning process can improve performance beyond vanilla prompting. However, existing approaches still face substantial scalability challenges. Although the latter achieves strong accuracy on dense and complex graphs, it requires assigning one agent per node, resulting in severe scalability issues and computational bottlenecks. Such coordination costs grow quickly with graph size, making the framework expensive for large instances. Meanwhile, real-world graphs continue to grow in both scale and complexity, often exceeding the effective reasoning capacity of a single LLM or a naively coordinated multi-agent system. Even as LLMs improve, larger and denser graphs will continue to pose new scalability challenges. This raises two key questions: (1) how can we enable effective reasoning over graphs that exceed current scalability limits, and (2) how can we obtain reliable answers without incurring prohibitive computational costs?

In this work, we revisit a classic principle of mathematics: divide the work, then integrate the answers back together. Inspired by distributed graph processing and the \textit{divide-and-conquer} paradigm~\cite{r21,r22}, we propose \ours, a multi-agent system that partitions nodes into subgraphs, assigns a dedicated agent to each subgraph, and employs a master agent to integrate outputs from each subgraph along with inter-subgraph information to solve the overall task for scalable graph algorithm reasoning. Specifically, we first cluster nodes into subgraphs and allocate an agent to each subgraph for intra-subgraph algorithm reasoning. The master agent then aggregates all outputs, incorporating inter-subgraph dependencies, to produce the final solution. Compared with node-level agent assignment, this design reduces the number of agents and coordination overhead while preserving the structural information needed for global reasoning. By constraining each agent to a manageable subgraph, the framework improves both scalability and reasoning reliability on complex graph instances.

We validate this framework through extensive experiments, demonstrating substantial gains in both accuracy and scalability. Our results show that \ours\ consistently outperforms strong baselines, with the largest improvements appearing on larger and denser graphs where single-agent methods degrade substantially. These findings indicate that subgraph-level decomposition is an effective way to scale LLM-based graph algorithm reasoning and provide a practical direction for handling graph instances beyond the range of current monolithic approaches.

Our main contributions are summarized as follows: 
\begin{itemize}
    \item To the best of our knowledge, we are the first to propose a divide-and-conquer multi-agent system (MAS) for scalable graph reasoning. Instead of assigning one agent to each node, our framework decomposes the original graph into subgraphs, enabling multiple agents to collaboratively solve large-scale graph reasoning problems.
    \item To support this divide-and-conquer MAS, we propose an effective graph decomposition algorithm tailored to downstream reasoning tasks. The proposed decomposition strategy produces subgraphs that are more suitable for later agent-level reasoning and facilitates the integration of local solutions into a global answer.
    \item Extensive experiments demonstrate the effectiveness and scalability of our framework. The results show that our method achieves strong performance on challenging graph reasoning tasks, particularly on larger and denser graphs where existing approaches face substantial difficulty.
\end{itemize}

The rest of the paper is organized as follows. Section~\ref{sec:related} reviews the related work. Section~\ref{sec:methods} presents the $\ours$ framework and describes how it is applied to graph reasoning tasks. Section~\ref{sec:experiments} evaluates the performance of $\ours$. Finally, Section~\ref{sec:conclusion} concludes the paper and discusses directions for future work.

\section{Related work}\label{sec:related}
We review three lines of research most relevant to our work: classical graph algorithms, machine learning-based graph reasoning methods, and recent LLM-based approaches for graph reasoning.

\subsection{Classical graph algorithms.}
Classical graph algorithms provide the foundation for reasoning over graph-structured data~\cite{van1990graph,even2011graph}. A large body of work has studied exact and efficient algorithms for fundamental tasks such as connectivity, shortest paths, cycle detection, and relational inference on graphs~\cite{tarjan1983data,tarjan1972depth,madkour2017survey}. These methods are often computationally well grounded and, for many canonical problems, admit strong theoretical guarantees. However, they are typically designed for structured graph inputs with explicitly defined nodes, edges, and objectives, and are less suited to settings where graph reasoning must be performed jointly with rich contextual information. In many modern applications, graph instances are accompanied by heterogeneous node or edge attributes, textual descriptions, external knowledge, or natural-language queries~\cite{r23,r24,r44}. Classical algorithms generally do not provide a native mechanism for incorporating such rich data into the reasoning process. As a result, while they remain highly effective for well-specified combinatorial problems, they are less flexible for complex graph reasoning scenarios that require integrating graph structure with unstructured or semantically rich information.

\subsection{Machine learning-based graph reasoning.}
To overcome the rigidity of purely algorithmic methods, a large literature has explored combining graphs with machine learning, especially through graph neural networks, graph transformers, and other graph representation learning frameworks~\cite{r26,r27,r28,r29}. These methods can encode both graph topology and node or edge attributes, and have achieved strong empirical performance across a wide range of applications, including social networks, bioinformatics, and knowledge graphs~\cite{r6,r7,r44,r8,r9,r10}. Compared with traditional algorithms, they are better able to incorporate rich data and learn task-relevant patterns from supervision. Nevertheless, most of these approaches remain task-specific in both design and training. Their success often depends on carefully chosen architectures, sampling strategies, positional or structural encodings, loss functions, and downstream decoders tailored to a particular task. Moreover, applying them to a new reasoning problem typically requires retraining, redesign, or both. This limits their usability in settings where one seeks a more general-purpose reasoning framework that can transfer across graph tasks without extensive task-specific engineering.

\subsection{LLM-based graph reasoning.}
More recently, researchers have begun exploring large language models for graph reasoning. This line of work is motivated by the strong in-context learning, interpretability, and task generalization capabilities of LLMs across diverse reasoning problems~\cite{r11,r12,r13,r14,r15,liu2024time,datta2025improving}. Existing efforts include prompting-based methods that serialize graphs into text, multimodal methods that augment graph reasoning with visual representations such as GITA~\cite{r3}, and multi-agent frameworks that distribute reasoning across agents~\cite{r2}. Compared with classical graph algorithms and learned graph encoders, these methods offer a more unified interface for reasoning over graph problems without task-specific training. However, scalability remains a major bottleneck. For single-agent LLM methods, reasoning quality degrades rapidly as graph size and density increase, since the model must track more structural dependencies and longer reasoning chains within a limited context and reasoning budget. Multi-agent methods partially alleviate this issue, but existing designs can still incur substantial coordination overhead; for example, assigning one agent per node is computationally expensive and becomes impractical on large graphs~\cite{r2}. In contrast, our work focuses on scalable graph algorithm reasoning through subgraph-level divide-and-conquer, which reduces coordination cost while preserving the global information needed for final reasoning.

\section{Proposed Method}\label{sec:methods}

To overcome the limitations discussed in Section~\ref{sec:introduction} and Section~\ref{sec:related} and better leverage the capabilities of LLMs for graph reasoning, we propose a novel multi-agent collaboration framework, \ours, as illustrated in Figure~\ref{fig:figure1}. The overall design follows a simple but effective intuition: instead of forcing a single LLM to reason over the entire graph at once, we decompose the original problem into smaller subproblems that can be handled by multiple agents, and then coordinate their outputs to recover the final answer. In this way, \ours\ reduces the reasoning burden on each individual agent while preserving the global information needed for solving the original task.

Specifically, the framework consists of two stages: \textbf{Graph-Query Decomposition} and \textbf{Hierarchical Reasoning}. In the first stage, the input graph and query are decomposed into a collection of manageable subgraphs together with corresponding sub-queries. In the second stage, sub-agents perform local reasoning within their assigned subgraphs, and a master agent integrates their outputs along with inter-subgraph dependencies to produce the final answer. This two-stage design addresses the challenges of complexity, efficiency, and accuracy in a unified and scalable manner.

\subsection{Graph-Query Decomposition}

We first describe how the original graph reasoning problem is decomposed. In a standard graph reasoning setting, the model takes as input (i) a graph $G = (V, E)$, where $V$ and $E$ denote the sets of vertices and edges, respectively, and (ii) a task specification $\mathcal{Q}$, which defines the target reasoning objective. Depending on the application, $\mathcal{Q}$ may correspond to a decision problem, a structural query, or a graph algorithmic reasoning task that requires combining local and global graph information.

Given a graph-query pair $(G, \mathcal{Q})$, our goal is to partition $G$ into compact and appropriately sized subgraphs so that each subproblem remains within the reasoning capacity of an individual LLM agent. At the same time, the decomposition should preserve sufficient structural information to support the reconstruction of a global solution. Formally, we apply a graph splitter $\mathcal{S}$ to partition $G$ into $n$ subgraphs:
\begin{equation}
    \mathcal{S}(G) = \{g^{(i)}\}_{i=1}^{n}.
\end{equation}
where each subgraph $g^{(i)} = (v^{(i)}, e^{(i)})$ contains a subset of internal nodes $v^{(i)}$ and internal edges $e^{(i)}$. In addition to the internal structure of each subgraph, we explicitly identify its \emph{exit nodes}, denoted by $g^{(i)}_{\text{exit}}$, namely, the nodes that connect to nodes in other subgraphs. These exit nodes play an important role in preserving cross-subgraph dependencies, since they act as the interface between local reasoning and global integration.

\begin{table}[t]
    \centering
    \caption{List of notations}
    \label{tab:notation}
    \scalebox{1}{
    \begin{tabular}{c|p{0.66\linewidth}}
        \toprule
        Notation & Description \\
        \midrule
        $G=(V,E)$ & Graph with nodes $V$ and edges $E$ \\
        $n$ & Number of subgraphs / sub-agents \\
        $\mathcal{Q}$ & Graph reasoning query \\
        $\mathcal{A}$ & Final answer \\
        $\mathcal{S}$ & Graph splitter \\

        $\mathcal{P}$ & Prompt template \\
        $\mathcal{D}$ & Question describer \\
        $\mathcal{R}$ & LLM reasoner \\
        $\mathcal{E}$ & Extractor \\
        $E_{\text{inter}}$ & Inter-subgraph edges \\  
        $g^{(i)}$ & The $i$-th subgraph \\
        $g_{\text{exit}}^{(i)}$ & Exit nodes of the $i$-th subgraph \\
        $q^{(i)}$ & Sub-query for the $i$-th subgraph \\
        $a^{(i)}$ & Raw sub-agent response \\
        $a_{\text{extract}}^{(i)}$ & Extracted sub-answer from the $i$-th sub-agent \\
        \bottomrule
    \end{tabular}
    }
\end{table}

Specifically, to avoid introducing bias from complicated partitioning methods and to keep the decomposition procedure general, we adopt a simple partitioning strategy as the splitter $\mathcal{S}$~\cite{newman2006modularity}. As shown later in the experiments, even such a lightweight decomposition strategy can provide substantial benefits when combined with hierarchical multi-agent reasoning. Besides, the splitter is actually method-agnostic to the specific strategy, and therefore can also be extended to more sophisticated decomposition strategies, which may further improve performance.

After the graph is partitioned, a question describer generates a task-aware sub-query for each subgraph. Specifically, for each $g^{(i)}$, the describer conditions on the exit nodes of the subgraph, the global task specification, and a prompt template for the reasoning task, producing a sub-query $q^{(i)}$:
\begin{equation}
    q^{(i)}=\mathcal{D}(g^{(i)}_{\text{exit}},\mathcal{P};\mathcal{Q})
\end{equation}
where $\mathcal{D}$ denotes the describer function and $\mathcal{P}$ is the prompt template. The purpose of the describer is to translate the original task $\mathcal{Q}$ into a localized reasoning objective for each subgraph, while still retaining awareness of how the local result may contribute to the final global answer. In other words, the sub-query is not simply a truncated version of the original question; it is a task-conditioned local reasoning instruction that accounts for the role of the subgraph in the broader graph structure.

This decomposition process produces a set of subgraph--subquery pairs $(g^{(i)}, q^{(i)})$ together with the set of inter-subgraph edges $E_{\text{inter}}$. Their relationship to the original graph can be written as
\begin{equation}
    G=\sum_{i} g^{(i)} + E_{\text{inter}},
\end{equation}
where $\sum_i g^{(i)}$ denotes the collection of all subgraphs and $E_{\text{inter}}$ captures the structural dependencies across them. The resulting decomposition, i.e., the set of local reasoning units $(g^{(i)}, q^{(i)})$ and the inter-subgraph connectivity information $E_{\text{inter}}$, is then passed to the next stage for hierarchical reasoning.

\subsection{Hierarchical Reasoning}

Given the decomposition results, \ours\ next performs reasoning in a hierarchical manner. This stage consists of two steps: (1) \emph{intra-subgraph local reasoning} by sub-agents, and (2) \emph{inter-subgraph global synthesis} by the master agent. Specifically, the intuition is that local agents focus on solving bounded reasoning problems within subgraphs, while the master agent focuses on combining local evidence into a globally consistent solution.

\subsubsection{Intra-subgraph reasoning.}
Each sub-agent is assigned one subgraph-subquery pair $(g^{(i)}, q^{(i)})$. The agent then applies the LLM reasoner $\mathcal{R}$ to generate a natural-language response:
\begin{equation}
    a^{(i)} = \mathcal{R}(g^{(i)},q^{(i)}).
\end{equation}

This local reasoning step allows each agent to concentrate on a restricted portion of the graph, thereby reducing the effective reasoning complexity. Since the input size is smaller and the structural scope is more limited, the agent can devote more capacity to the actual reasoning process rather than spending its context budget on representing an overly large graph.

However, the raw output $a^{(i)}$ of an LLM is often verbose and may include intermediate reasoning steps, explanatory text, or redundant details that are unnecessary for downstream aggregation. Passing these full responses directly to the master agent would increase the integration burden and may introduce noise. To address this issue, we employ an Extractor $\mathcal{E}$ that distills each response into a concise sub-answer, denoted by $a^{(i)}_{\text{extract}}$. The extractor retains only the essential information needed for final reasoning, such as local conclusions, critical structural facts, or task-relevant summaries. This refinement step improves communication efficiency between agents and helps the master agent focus on the information that matters most.

\subsubsection{Inter-subgraph synthesis.}
With the sub-answers from sub-agents, the master agent performs global synthesis. Specifically, it aggregates the set of extracted sub-answers $\{a^{(i)}_{\text{extract}}\}$ together with the inter-subgraph information $E_{\text{inter}}$, and then uses the same reasoning module $\mathcal{R}$ to generate the final answer $\mathcal{A}$ to the original query $\mathcal{Q}$:
\begin{equation}
    \mathcal{A}=\mathcal{R}(\sum_{i}a_{\text{extract}}^{(i)},E_{\text{inter}};\mathcal{Q}).
\end{equation}

Note that the role of the master agent is not to redo all reasoning from scratch, but to integrate local evidence into a coherent global solution. It resolves dependencies that span multiple subgraphs, reconciles potentially complementary local observations, and determines how inter-subgraph edges affect the final answer. This design allows the overall system to preserve a global reasoning view without requiring any single agent to directly process the full input graph.

Overall, the proposed hierarchical reasoning framework transforms a difficult monolithic graph reasoning problem into a coordinated multi-agent process with clear functional roles. Sub-agents are responsible for localized inference, while the master agent is responsible for structured global integration. As we will show later in Section~\ref{sec:experiments}, by decomposing both the graph and the reasoning process, \ours\ improves scalability while maintaining the ability to solve graph reasoning tasks that require global consistency.

We summarize the overall procedure of \ours\ in Algorithm~\ref{alg:code}. Lines 1-5 correspond to the \textbf{Graph-Query Decomposition} stage, where the input graph is partitioned into subgraphs, the inter-subgraph connections are identified, and a task-aware sub-query is generated for each subgraph. Lines 6-12 implement the \textbf{Hierarchical Reasoning} stage. In this stage, each sub-agent performs local reasoning on its assigned subgraph, an extractor distills the local response into a concise sub-answer, and a master agent integrates all extracted sub-answers with inter-subgraph information to produce the final answer to the original query.

\begin{algorithm}[t]
\caption{The \ours\ Framework}
\label{alg:code}
    \begin{algorithmic}[1]
    \REQUIRE Graph-query pair $(G,\mathcal{Q})$, prompt template $\mathcal{P}$, graph splitter $\mathcal{S}$, question describer $\mathcal{D}$, LLM reasoner $\mathcal{R}$, and extractor $\mathcal{E}$
    \ENSURE Final answer $\mathcal{A}$
    \STATE Partition the input graph into subgraphs: $\mathcal{S}(G)=\{g^{(i)}\}_{i=1}^{n}$
    \STATE Identify the inter-subgraph edges $E_{\text{inter}}$
    \FOR{$i=1$ to $n$}
        \STATE Generate the task-aware sub-query: $q^{(i)}=\mathcal{D}(g^{(i)}_{\text{exit}},\mathcal{P};\mathcal{Q})$
    \ENDFOR
    \FOR{$i=1$ to $n$}
        \STATE Perform intra-subgraph reasoning: $a^{(i)}=\mathcal{R}(g^{(i)},q^{(i)})$
        \STATE Extract the concise sub-answer: $a_{\text{extract}}^{(i)}=\mathcal{E}(a^{(i)})$
    \ENDFOR
    \STATE Perform inter-subgraph synthesis:
    \STATE \hspace{1em} $\mathcal{A}=\mathcal{R}\!\left(\sum_{i} a_{\text{extract}}^{(i)},E_{\text{inter}};\mathcal{Q}\right)$
    \STATE \textbf{Return} $\mathcal{A}$
    \end{algorithmic}
\end{algorithm}

\section{Experiments}\label{sec:experiments}

\begin{table*}[t]
\centering
\caption{Dataset statistics}
\scalebox{0.95}{
\begin{tabular}{lcccccc}
\toprule
\multirow{3}{*}{Task} &
\multicolumn{3}{c}{NLGraph~\cite{r}} &
\multicolumn{3}{c}{Ours} \\
\cmidrule(lr){2-4}\cmidrule(lr){5-7}
 & Graphs & Nodes & Edges & Graphs & Nodes & Edges \\
 & & (nodes per graph) &(edges per graph) & &(nodes per graph) & (edges per graph) \\
\midrule
Connectivity & 224 & 4,651 (20.76) & 16k (71.42) & 3,000 & 151k (50.33) & 221k (73.67) \\
Cycle        & 1,150 & 24k (20.87) & 24k (20.86) & 3,000 & 150k (50) & 271k (90.33) \\
Shortest path & 380 & 4,450 (11.71) &  8,342 (21.95) & 1,500 &  76k (50.66) & 197k (131.33) \\
\bottomrule
\end{tabular}
}
\label{tab:dataset}
\end{table*}

In this section, we will answer the following research questions:
\begin{itemize}
    \item \textbf{Question 1:} Does \ours\ outperform existing LLM-based baselines on graph algorithm reasoning tasks?
    \item \textbf{Question 2:} How do existing methods behave as graph size increases?
    \item \textbf{Question 3:} Does \ours\ provide larger gains on more challenging and larger graphs?
\end{itemize}

\subsection{Setup}\label{sec:datasets}
We compared our method with existing methods on two datasets: one dataset is from the NLGraph~\cite{r} paper, and another one was constructed by us that still follows the same graph generation procedure as NLGraph but increases the size of the graphs.

The reason we create and evaluate on a new dataset constructed by us is that existing datasets for graph algorithm reasoning mostly focus on relatively small graphs, which are easier for current LLM-based methods to process and reason over~\cite{r,r1}. In contrast, larger graphs with more nodes and denser structural dependencies are substantially more challenging, yet remain underrepresented in current benchmarks. This imbalance makes it difficult to systematically evaluate how graph reasoning methods scale as graph size increases.

Specifically, the dataset we constructed contains an equal number of graphs in each size range: 0-20, 20-40, 40-60, 60-80, and 80-100 nodes. This design allows us to study model behavior under progressively more difficult graph reasoning settings instead of concentrating most evaluations on small graphs. In addition, for the shortest path and connectivity tasks, we deliberately select target node pairs that are relatively far apart. This makes the reasoning process more demanding, since the model must track longer-range graph dependencies and perform more nontrivial multi-step reasoning. As a result, task difficulty naturally increases with graph size, making the benchmark more suitable for assessing scalability.

Table~\ref{tab:dataset} summarizes the statistics of the two datasets. Compared with NLGraph~\cite{r}, our benchmark contains substantially more graphs, nodes, and edges, and provides a more challenging testbed for evaluating graph algorithm reasoning methods, especially in large-graph settings where scalability becomes critical.

\subsection{Experiment Results}
We first evaluate on the original NLGraph dataset as in Table~\ref{tab:results_on_NLGraph}, following the experimental setting of NLGraph~\cite{r}. We compare against the results reported in the NLGraph for Few-Shot~\cite{r31} and Chain-of-Thought (CoT)~\cite{r30} baselines. We also evaluate \ours on the dataset we constructed as described in subsection~\ref{sec:datasets} and compare it with the same representative LLM-based baselines as in Table~\ref{tab:results}, also using Few-Shot~\cite{r31} and Chain-of-Thought (CoT)~\cite{r30} baselines following the experimental setting of NLGraph~\cite{r}.
All methods are implemented with GPT-4.1-mini without any fine-tuning. For NLGraph dataset as in Table~\ref{tab:results_on_NLGraph}, we report accuracy across three difficulty levels: easy, medium, and hard following their original paper. For our dataset as in Table~\ref{tab:results}, we report accuracy across five graph size ranges to better examine overall performance and how each method behaves as graph size increases. Code and dataset are available at \hyperlink{https://anonymous.4open.science/r/GraphDCA-FBDF}{https://anonymous.4open.science/r/GraphDCA-FBDF}.


\begin{table*}[t]
\caption{Accuracy on the NLGraph dataset, best results are bold.}
\label{tab:results_on_NLGraph}
\centering
\scalebox{1.0}{
\begin{tabular}{llccc}
\toprule
\multirow{2}{*}{Task} & \multirow{2}{*}{Method} & \multicolumn{3}{c}{Difficulty} \\
\cmidrule(lr){3-5}
& & easy & medium & hard \\
\midrule
\multirow{3}{*}{Connectivity} & NLGraph-FewShot~\cite{r31} & 93.8\% & 83.8\% & 76.6\% \\
& NLGraph-CoT~\cite{r30} & 94.3\% & 82.2\% & 77.2\% \\
& \ours & \textbf{94.8\%}(1\%$\uparrow$) & \textbf{90.7\%}(8\%$\uparrow$) & \textbf{83.1\%}(8\%$\uparrow$) \\
\midrule
\multirow{3}{*}{Shortest path} & NLGraph-FewShot~\cite{r31} & 49.2\% & --$^{*}$ & 35.7\% \\
& NLGraph-CoT~\cite{r30} & 76.8\% & --$^{*}$ & 35.8\% \\
& \ours & \textbf{97.7\%}(27\%$\uparrow$) & --$^{*}$ & \textbf{98.0\%}(174\%$\uparrow$) \\
\midrule
\multirow{3}{*}{Cycle} & NLGraph-FewShot~\cite{r31} & 80.0\% & 70.0\% & 61.0\% \\
& NLGraph-CoT~\cite{r30} & \textbf{84.7\%} & 63.3\% & 53.3\% \\
& \ours & 81.3\% & \textbf{81.5\%}(16\%$\uparrow$) & \textbf{80.2\%}(31\%$\uparrow$) \\
\bottomrule
\end{tabular}
}
\vspace{0.5em}
\\
{\raggedleft $^{*}$ Not provided in their original paper \hspace{4em} \par}
\end{table*}


\subsubsection{Q1: \ours\ Outperforms Existing Baselines.}
As shown in Table~\ref{tab:results_on_NLGraph} and Table~\ref{tab:results}, \ours\ consistently achieves higher accuracy than all baselines across nearly all tasks and graph size groups. This confirms the effectiveness of our divide-and-conquer multi-agent design, which distributes the reasoning load across specialized agents and reduces the burden on any single LLM. Specifically, we noticed that the performance improvements of \ours compared with baselines are larger on our datasets in Table~\ref{tab:results}, especially on larger graphs with sizes between 80 and 100; this demonstrates the advantage of our method on large graph reasoning compared with existing methods. For small graphs, single-agent methods remain competitive on connectivity and shortest-path tasks, and in some cases slightly outperform \ours\ on connectivity, likely because decomposition is unnecessary in such simple settings and may occasionally produce less informative subgraphs. However, as graph size increases, the performance of single-agent methods deteriorates quickly, particularly on cycle detection and shortest-path tasks, whereas \ours\ remains substantially more robust. These results indicate that decomposing the graph into manageable subproblems is effective for preserving reasoning quality in more complex settings.


\subsubsection{Q2: Existing Methods Struggle on Large Graphs.} 
As shown in Table~\ref{tab:results_on_NLGraph} and Table~\ref{tab:results}, existing methods have limited reasoning ability on larger graphs. Their accuracy drops sharply as graph size increases, and on harder tasks they often approach failure. Specifically, existing methods demonstrate the worst performance on our dataset with graph sizes larger than 80, as shown in Table~\ref{tab:results}, further demonstrating their struggle on large graphs. This trend is especially clear for shortest-path and cycle detection, both of which require multi-step reasoning over long-range structural dependencies. For example, in cycle detection with graphs larger than 40 nodes, the baselines often default to answering “yes” for most queries, yielding accuracy close to 50\% on the balanced test set.
This suggests that standard prompting and CoT prompting are insufficient for reliable reasoning once graph complexity exceeds the effective capacity of a single model. In contrast, \ours\ uses graph decomposition and coordinated subgraph reasoning to better manage this complexity, maintaining consistently higher accuracy even in challenging large-graph settings.

\subsubsection{Q3: \ours\ Performs Better on Large Graphs.} 
Table~\ref{tab:results_on_NLGraph} and Table~\ref{tab:results} also report the relative improvement of \ours\ over single-agent baselines. We observe that the gains become larger as the task becomes more difficult, indicating that our framework is particularly effective in challenging reasoning scenarios. In particular, for hard graphs in the NLGraph dataset, as in Table~\ref{tab:results_on_NLGraph}, and graphs with more than 60 nodes in our dataset, as in Table~\ref{tab:results}, \ours\ achieves nearly twofold accuracy gains on the shortest-path task, highlighting its advantage on problems that require deeper and more global reasoning. Even on connectivity, where all methods perform strongly on small graphs, the benefit of \ours\ becomes increasingly visible as graph size grows. Overall, these results demonstrate that the proposed framework scales more effectively than existing baselines and is better suited for graph reasoning on large and complex instances.

\begin{table*}[t]
  \caption{Accuracy on our dataset, best results are bold.}
  \label{tab:results}
  \centering
  \scalebox{0.85}{
  \begin{tabular}{llccccc}
    \toprule
    \multirow{2}{*}{Task} & \multirow{2}{*}{Method} & \multicolumn{5}{c}{Graph size} \\
    \cmidrule(lr){3-7}
    & & 0-20 & 20-40 & 40-60 & 60-80 & 80-100  \\
    \midrule
            & NLGraph-FewShot~\cite{r31} & 99.8\% & 99.5\% & 72.5\% & 74\% & 69.5\%  \\
    Connectivity & NLGraph-CoT~\cite{r30} & \textbf{100\%} & \textbf{99.7\%} & 72.3\% & 70.5\% & 66.0\%  \\
            & \ours   & 99.8\% & \textbf{99.7\%} & \textbf{85.5\%}(18\%$\uparrow$) & \textbf{78.5\%}(11\%$\uparrow$) & \textbf{81.7\%}(24\%$\uparrow$) \\
    \midrule
             & NLGraph-FewShot~\cite{r31} & 88.7\% & 58.3\% & 27.0\% & 18.7\% & 16.0\% \\
    Shortest path & NLGraph-CoT~\cite{r30} & 89.0\% & 57.3\% & 27.6\% & 16.0\% & 18.0\% \\
             & \ours   & \textbf{91.7\%}(3\%$\uparrow$) & \textbf{86.7\%}(49\%$\uparrow$) & \textbf{65.7\%}(143\%$\uparrow$) &\textbf{54.3\%}(190\%$\uparrow$) & \textbf{44.3\%}(177\%$\uparrow$) \\
    \midrule
          & NLGraph-FewShot~\cite{r31} & 61.7\% & 53.8\% & 50.2\% & 50\% & 50\%  \\
    Cycle & NLGraph-CoT~\cite{r30} & 59.7\% & 51.2\% & 50\% & 50\% & 50\%  \\
          & \ours   & \textbf{85.8\%}(39\%$\uparrow$) & \textbf{81.3\%}(51\%$\uparrow$) & \textbf{66.0\%}(31\%$\uparrow$) & \textbf{58.8\%}(18\%$\uparrow$) & \textbf{52.6\%}(5\%$\uparrow$) \\
    \bottomrule
  \end{tabular}
  }
\end{table*}


\subsection{Case study}

We present a case study to illustrate how \ours\ improves connectivity reasoning on larger graphs. Consider a graph with 100 nodes that can be naturally divided into two clusters, denoted by cluster 1 and 2. Most edges lie within each cluster, while only a few sparse edges connect the two clusters. Suppose the task is to determine whether a node in cluster 1 (specifically, the node 27) is connected to another in cluster 2 (node 97).

When the entire graph is given to a single LLM agent, the model must reason over a long path spanning two clusters and identify the few critical cross-cluster edges that connect them. This is difficult because the relevant evidence is distributed across distant parts of the graph: the model must first verify that if the node 27 can reach some exit node in cluster 1, then determine whether that exit node is linked to an exit node in cluster 2, and finally check whether the node 97 is reachable from that entry point inside cluster 2. In practice, a single agent may miss one of these steps or focus excessively on local structure, leading to an incorrect prediction of disconnection.

In contrast, \ours\ first decomposes the graph into two clusters and assigns a sub-agent to each cluster. The sub-agent for cluster 1 only needs to determine whether the node 27 can reach an exit node of cluster 1, while the sub-agent for cluster 2 analogously checks whether an exit node of cluster 2 can reach the node 97. Each subproblem is substantially simpler than reasoning over the entire graph, since it is restricted to a smaller subgraph with fewer irrelevant nodes and edges.

\begin{figure}[t]
    \centering
    \includegraphics[width=\linewidth]{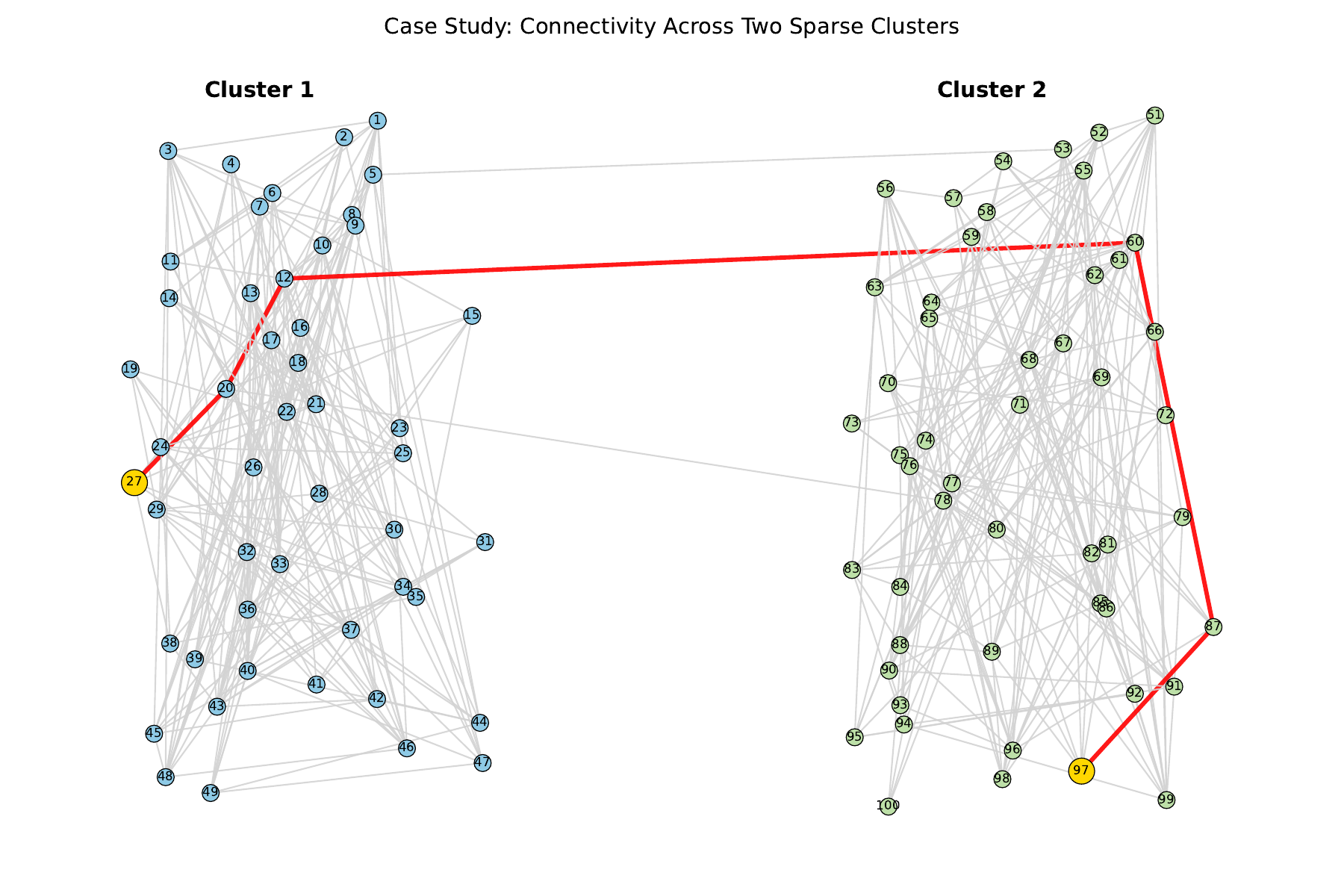}
    \caption{Case study of connectivity reasoning with \ours. A single-agent approach must reason over the entire graph to determine whether node 27 and node 97 are connected, requiring it to track a long-range path across multiple clusters. In contrast, \ours\ decomposes the graph into two clusters and assigns one sub-agent to each cluster for local reachability reasoning. The master agent then integrates the local results with the sparse inter-cluster edges to infer the final global connectivity.}
    \label{fig:casestudy}
    \vspace{-1.25em}
\end{figure}

The master agent then combines the outputs of the two sub-agents corresponding to the two clusters with the sparse inter-cluster edges. If an exit node reachable from the node 27 in cluster 1 is connected via a cross-cluster edge to an exit node that can reach node 97 in cluster 2, the master agent concludes that the node 27 and the node 97 are connected in the original graph through the exit nodes in both clusters. In this way, \ours\ recovers the global connectivity result by composing two local reachability decisions with a small amount of cross-cluster information.

This example illustrates the core advantage of our framework. Rather than forcing a single LLM to track a long-range dependency across the entire graph, \ours\ decomposes the task into two local reasoning steps followed by one lightweight global aggregation step. This divide-and-conquer process makes the reasoning problem more tractable and improves the reliability of the final connectivity prediction.

\section{Conclusion}\label{sec:conclusion}

In this work, we proposed \ours, a divide-and-conquer multi-agent framework for scalable graph algorithm reasoning. Instead of relying on a single model to process the entire graph, our approach decomposes the input graph into manageable subgraphs, assigns each subgraph to an individual agent for local reasoning, and then uses a master agent to integrate the local outputs and inter-subgraph information to answer the original question. In this way, \ours\ reduces the reasoning burden on each agent while preserving the global structural information required for solving the task. Experimental results demonstrate that \ours\ consistently outperforms existing methods on large-graph algorithm reasoning tasks, showing clear advantages especially as graph size and task complexity increase. These findings highlight both the effectiveness and the scalability of our framework, and suggest that structured multi-agent collaboration is a promising direction for graph reasoning with LLMs. In future work, it would be interesting to explore more advanced graph decomposition strategies and extend the framework to a broader range of graph reasoning tasks.


\clearpage
\newpage

\bibliographystyle{abbrv}
\bibliography{reference.bib}

\newpage
\appendix
\section{Appendix}
\begin{strip}
\centering
\captionof{table}{Accuracy on the Triangle task, best results are bold.}
\label{tab:results_triangle}
\scalebox{1.0}{
\begin{tabular}{llccccc}
  \toprule
  \multirow{2}{*}{Task} & \multirow{2}{*}{Method} & \multicolumn{5}{c}{Graph size} \\
  \cmidrule(lr){3-7}
  & & 0-20 & 20-40 & 40-60 & 60-80 & 80-100 \\
  \midrule
  Triangle & Graphwiz~\cite{r1} & \textbf{86.0\%} & 72.5\% & 64.0\% & 35.0\% & 42.5\% \\
           & \ours & 83.0\% & \textbf{82.5\%}(14\%$\uparrow$) & \textbf{79.5\%}(24\%$\uparrow$) & \textbf{62.5\%}(79\%$\uparrow$) & \textbf{73.0\%}(72\%$\uparrow$) \\
  \bottomrule
\end{tabular}
}
\end{strip}
Besides the conventional graph reasoning tasks of connectivity, cycle detection, and shortest path introduced above, we additionally design a Triangle benchmark following Graphwiz~\cite{r1}. Like NLGraph~\cite{r}, Graphwiz primarily evaluates relatively small graphs, which are comparatively easier for current LLM-based methods to process and reason over. To obtain a more balanced evaluation across graph sizes, we follow the same graph generation procedure as Graphwiz~\cite{r1}, but adjust the sampling distribution so that the dataset contains an equal number of graphs in each size range: 0-20, 20-40, 40-60, 60-80, and 80-100 nodes. For this supplementary evaluation, we report the performance of \ours using GPT-4.1-mini without fine-tuning under the Graphwiz protocol. This construction makes the Triangle benchmark better suited for analyzing how graph reasoning performance scales with graph size.

As shown in Table~\ref{tab:results_triangle}, \ours is competitive with Graphwiz on the smallest graphs and consistently performs better on medium and large graphs, with the gains becoming more pronounced as graph size increases. This indicates that the proposed framework is effective on the Triangle task, especially on harder instances.

\end{document}